\PassOptionsToPackage{table,dvipsnames}{xcolor}
\documentclass{article} 
\usepackage[T1]{fontenc}
\usepackage{iclr2027_conference,times}
\iclrfinalcopy

\usepackage{amsmath,amsfonts,bm}

\def\eqref#1{equation~\ref{#1}}
\def\Eqref#1{Equation~\ref{#1}}

\def\1{\bm{1}}

\DeclareMathAlphabet{\mathsfit}{\encodingdefault}{\sfdefault}{m}{sl}
\SetMathAlphabet{\mathsfit}{bold}{\encodingdefault}{\sfdefault}{bx}{n}

\usepackage{hyperref}
\usepackage{url}
\usepackage[table,dvipsnames]{xcolor}
\usepackage{xspace}
\usepackage{algorithm}
\usepackage{algpseudocode}
\algrenewcommand{\algorithmicensure}{\textbf{Output:}}
\usepackage{booktabs}
\usepackage{graphicx}
\usepackage{multirow}
\usepackage{makecell}
\usepackage{tikz}
\usepackage{fontawesome5}
\usetikzlibrary{arrows.meta,positioning,fit,backgrounds}
\usepackage{pifont}
\definecolor{turbo}{HTML}{E0781E}
\definecolor{metablue}{HTML}{3F6FB0}
\definecolor{harnessgray}{HTML}{8A9199}

\definecolor{darkblue}{rgb}{0, 0, 0.5}
\definecolor{oursrow}{rgb}{0.88, 0.92, 1}
\hypersetup{colorlinks=true, citecolor=darkblue, linkcolor=darkblue, urlcolor=darkblue}
\usepackage{enumitem}

\usepackage[most]{tcolorbox}
\tcbuselibrary{listings}
\definecolor{codeback}{HTML}{F6F8FA}
\definecolor{codeframe}{HTML}{D0D7DE}
\newtcblisting{codebox}{
  listing only,
  colback=codeback,
  colframe=codeframe,
  boxrule=0.5pt,
  arc=2pt,
  left=5pt, right=4pt, top=2pt, bottom=2pt,
  listing options={
    basicstyle=\footnotesize\ttfamily,
    columns=fullflexible,
    keepspaces=true,
    breaklines=false,
    upquote=true,
  },
}

\newcommand{\authorsep}[0]{\ \ }

\title{Turbo Harness:\\Instance-Adaptive Harness Optimization}

\author{\leavevmode\unskip
\textbf{Tunyu Zhang}\space$^{{*} , \text{\textdagger}, 1 , 2} $\authorsep
\textbf{Hao Wang}\space$^{{*},2, 3}$\authorsep
\textbf{Kai Xu}\space$^{2, 3}$\authorsep
\textbf{Dimitris N. Metaxas}\space\space$^{1}$\authorsep
\\
$^1$ Rutgers University \authorsep
$^2$ Red Hat AI Innovation \authorsep
$^3$ MIT-IBM Watson AI Lab
}

\begin{document}

\maketitle

\begingroup
\makeatletter
\long\def\@makefntext#1{%
  \setlength{\parindent}{0pt}%
  \noindent#1%
}
\makeatother
\renewcommand{\thefootnote}{}

\footnotetext[0]{%
 $^{*}$Equal contribution.  $^{\text{\textdagger}}$Work is done during an internship at the Red Hat AI innovation team. Correspondence to:
  Tunyu Zhang (\nolinkurl{ty.zhang@rutgers.edu}), 
  Dimitris N. Metaxas (\nolinkurl{dnm@cs.rutgers.edu}).
}
\endgroup

\begin{abstract}
Automating the search for effective harnesses is an important step toward enabling agents to recursively self-improve. Existing harness optimizations typically produce a single global harness that is applied uniformly across task instances. However, a harness that works well on average may not be optimal for every instance. We introduce \emph{Turbo Harness}, a framework that can adapt a globally optimized harness to each instance by reusing information generated during the original optimization process. Specifically, Turbo Harness recycles artifacts produced during a completed global harness optimization run, and summarizes them into a structured playbook. We train a harness editor to leverage this prior optimization experience to generate instance-specific patches to the global harness. At inference time, the editor uses the instance and the playbook to construct a tailored harness in which the execution model operates. Through numerical experiments, we show that Turbo Harness consistently outperforms existing harness optimization baselines across seven benchmarks spanning interactive agent tasks, software engineering, and long-horizon terminal tasks. 

\smallskip
\noindent\textbf{\faGithub\ Code:}
\href{https://github.com/Tyrion58/turbo-harness}
{\texttt{github.com/Tyrion58/turbo-harness}}
\end{abstract}

\section{Introduction}

Large language models (LLMs) and their surrounding harnesses are two fundamental building blocks of modern agents \citep{tian2026swe}. The notion of a harness has expanded well beyond simply exposing an LLM to an environment or a collection of tools~\citep{yao2022react}. Today, it encompasses nearly all of the runtime machinery surrounding the model (e.g., context management, memory, workflow orchestration, and verification)~\citep{huang2026memoharness, zhong2026ai, zhou2026externalization}. These mechanisms can substantially extend the effective capabilities of a static LLM, particularly on long-horizon, scientific, and open-ended tasks~\citep{ma2026longhorizon, wu2026structagent}. At the same time, this broad definition creates an enormous design space: for an application of interest, one must decide what information to expose to the model, what state to maintain, when and how to intervene, and how much computation to allocate. Designing an effective harness is therefore a challenging optimization problem in its own right.

A growing line of work seeks to automate this process through \emph{harness optimization} \citep{young2025effective, lee2026meta, lee2026recursive}. For example, Meta-Harness~\citep{lee2026meta} adopts an outer-loop program-search approach, in which a separate agent proposes new harnesses based on the execution traces of previously evaluated candidates; Harness-R1~\citep{shao2026harness} trains a harness engineer with reinforcement learning (RL) to transform batches of agent failures into executable runtime patches. 
Despite their promise, existing approaches have two limitations. First, they optimize a single global harness that is then applied uniformly across all, or a batch of, test instances. Yet a harness that is effective for one instance may be suboptimal for another. For example, SWE-Bench-style instances may come from different GitHub repositories with distinct development workflows and testing conventions, and may benefit from different harness configurations. 
Second, harness optimization is expensive and much of the information generated during this process is ultimately underutilized. During optimization, the system accumulates a rich history of candidate harnesses and evidence about which strategies succeed or fail under different conditions. Existing methods use this experience to improve the global harness \citep{lee2026meta, lee2026recursive, sengupta2026harbor}. Once optimization is complete, however, much of the instance-specific signal contained in this history is discarded. 
This leads to a fundamental question:
\begin{center}
\emph{Can we adapt a globally optimized harness to each individual instance by reusing the experience accumulated during harness optimization?}
\end{center}

\begin{figure}[t]
\centering
\includegraphics[width=\linewidth]{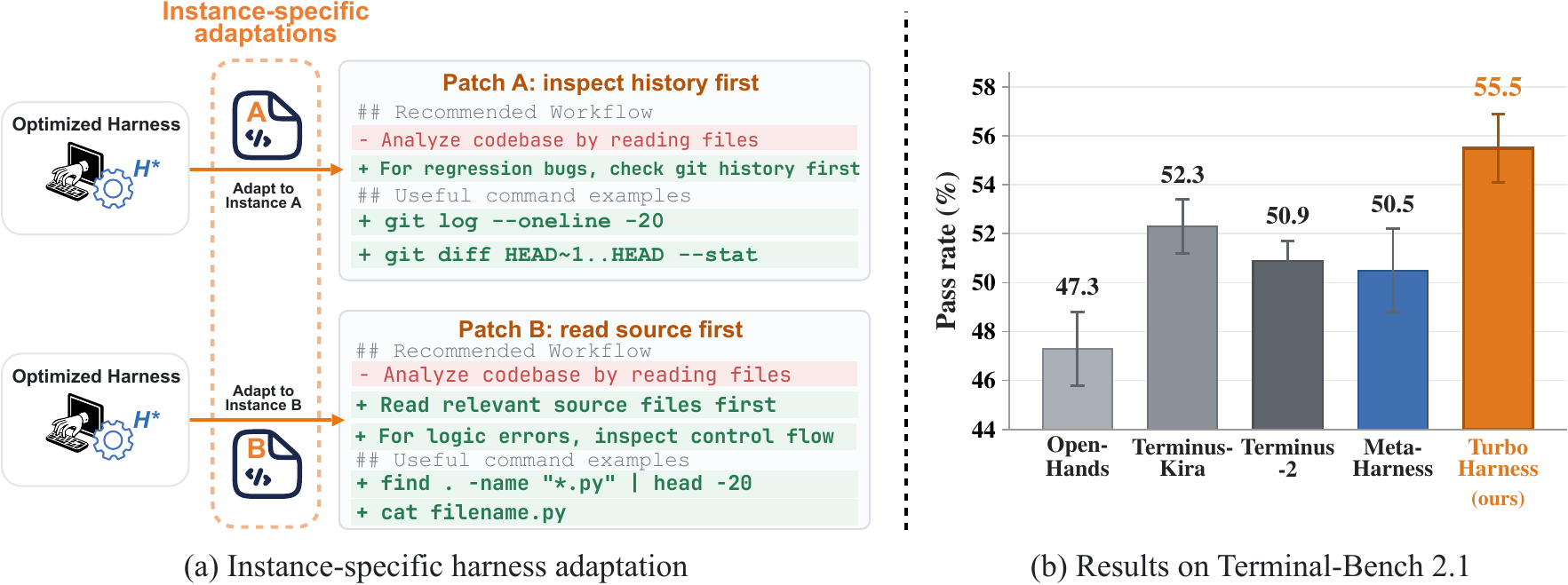}
\vspace{-2em}
\caption{\textbf{Turbo Harness adapts a globally optimized harness $\texttt{H}^\star$ to each instance.} \emph{(a)}~A harness editor produces an instance-specific patch to $\texttt{H}^\star$, yielding a per-instance harness. \emph{(b)}~On Terminal-Bench-2.1 (Claude Sonnet 4.5, $44$-task test split), Turbo Harness achieves the highest pass rate compared with several widely-used human-designed harnesses.}
\label{fig:teaser}
\vspace{-1em}
\end{figure}

In this paper, we introduce \textbf{\emph{Turbo Harness}}\footnote{We adopt the name ``Turbo Harness'' by analogy to a turbocharger, which recovers energy from exhaust gas and reuses it to improve engine performance.}. It is a new harness optimization framework that reuses artifacts from a completed outer-loop harness search to adapt a globally optimized harness to individual task instances. 

Specifically, we reuse the search archive produced during outer-loop optimization and summarize it into a structured playbook \citep{zhang2026agentic}. This playbook records both successful and unsuccessful harness-editing strategies, including the specific edits that lead to performance improvements and those that fail to do so. 
We then train a \emph{harness editor}, an RL-fine-tuned open-source model responsible for adapting the global harness to each instance. At deployment time, the harness editor receives a test instance together with access to the playbook and proposes a patch to the globally optimized harness. The resulting instance-specific harness is then used by the main execution model, which can be any model, to solve the test instance. Figure~\ref{fig:teaser} illustrates this instance-specific adaptation and shows its gains on Terminal-Bench-2.1.

Our framework offers three advantages. First, instance-adaptive harnessing provides more flexibility than a single global harness across all task instances, consistently improving task performance in our experiments. For example, on SWE-smith-MR, Turbo Harness improves pass rate from 50.7\% to 64.0\% with Claude Haiku 4.5 and from 70.7\% to 88.0\% with Gemini 3.7 Flash. 
Second, training the harness editor is lightweight: rather than generating a new harness from scratch, it only needs to patch a globally optimized harness, guided by a playbook that summarizes successful and unsuccessful edits from the outer-loop search. This allows us to use a relatively small editor model (Qwen3.5-9B in our experiments). Finally, Turbo Harness adds little inference-time overhead because the editor is called only once per instance; moreover, the adapted harness can often reduce the number of execution steps required by the much larger execution model, potentially even lowering overall inference-time compute (see e.g., Table~\ref{tab:coding} and \ref{tab:tb21}).

In short, our contributions are:
\begin{itemize}[leftmargin=2em]
\item We formulate \emph{instance-adaptive harness optimization} and highlight the limitations of deploying a single globally optimized harness across all task instances.
\item We propose Turbo Harness, which reuses the artifacts collected during the outer-loop harness search to train a harness editor that adapts the global harness to each individual task instance.
\item We show that Turbo Harness is lightweight: the editor only patches a globally optimized harness, can be instantiated with a small model, and is invoked only once per task instance.
%
\item We demonstrate that instance-specific harness adaptation consistently outperforms globally optimized harnesses across seven benchmarks spanning interactive agent tasks, software engineering, and long-horizon terminal tasks, while often reducing execution steps and costs.
\end{itemize}

\section{Related Work}
\textbf{Prompt and Context Optimization.}\quad
A growing line of work improves frozen language models by optimizing their prompts or, more broadly, the information provided to them during inference~\citep{ramnath2025systematic, li2025survey}. 
For example, GEPA~\citep{agrawal2026gepa} evaluates execution trajectories to evolve prompt candidates via a Pareto-based search. ACE~\citep{zhang2026agentic} accumulates and refines an evolving playbook of strategies through generation, reflection, and curation. MCE~\citep{ye2026meta} further expands this design space by co-evolving context-engineering skills and query-conditioned context functions, whose artifacts may include files and code. They primarily optimize the model's context. Harness engineering, however, encompasses a broader design space: beyond deciding what information is presented to the model, it can modify the executable runtime itself, including control flow, tool-use workflows, permissions, and verification mechanisms.

\textbf{Automatic Harness Optimization.}\quad
Recent methods~\citep{sengupta2026harbor, lin2026agentic, pan2026evolving, park2026autosaddler} optimize the agent scaffold itself. Meta-Harness~\citep{lee2026meta} uses a strong external coding agent to search over executable harnesses by inspecting previous implementations, scores, and execution traces. Self-Harness~\citep{zhang2026self} removes the stronger external proposer, allowing the target model to identify failure patterns, propose harness modifications, and validate them through testing. Both methods produce harnesses intended to generalize across a collection of instances. AutoDesign~\citep{luo2026autodesign} likewise applies meta-harness optimization to long-horizon agentic design tasks. Recursive Harness Self-Improvement~\citep{lee2026recursive} performs lightweight optimization by representing the agent loop as a prompt-level object and repeatedly revising it using pairwise feedback from its revision history. Unlike them, our method introduces an instance-adaptive harness without requiring a new multi-round search for each instance. This strategy enables more flexibility, which translates into consistent performance gains in our experiments.

\textbf{Adaptive Prompt and Harness Optimization.}\quad
A related line of work studies prompt and harness optimization that tailors the resulting prompt or harness to individual task instances.
For example, Advisor Models~\citep{asawa2025train} trains an advisor model that enables instance-specific prompt optimization. TTHE~\citep{nie2026tthe} performs harness search on unlabeled test batches and carries the selected program to subsequent batches. Harness-R1~\citep{shao2026harness} trains a harness engineer from batches of failure trajectories and evaluates each candidate patch by rerunning the same tasks under the modified harness. A concurrent work JIT-Agent~\citep{zhang2026jit} also targets instance-specific harnesses, but generates each harness from scratch via teacher distillation and evolutionary RL. In contrast, Turbo Harness at deployment time introduces only a single additional call to a small model, avoiding the two-pass procedure required by Harness-R1 \citep{shao2026harness} to first collect a failure trajectory and then rerun the task under the resulting patch. Moreover, Turbo Harness edits an existing global harness rather than synthesizing a new harness from scratch for each task instance.

\section{Preliminaries: Harness Optimization}
\label{sec:preliminaries}



\textbf{Objective of Harness Optimization.}\quad
An agent consists of a language model $\texttt{M}$ and a harness $\texttt{H}$. Given a task instance $x$, the agent executes $\texttt{M}$ under $\texttt{H}$ to interact with the task environment and produce an output. The harness is an executable program that governs this interaction, including context construction, tool use, state management, and execution control. 

Given a frozen model $\texttt{M}$, a task distribution $\mathcal{T}$, and an admissible harness space $\mathcal{H}$, harness optimization seeks a harness that maximizes the expected task performance:
\begin{equation}
  \texttt{H}^\star \in \arg\max_{\texttt{H} \in \mathcal{H}}
  \mathbb{E}_{x \sim \mathcal{T}}\left[r(\mathsf{Agent}(\texttt{M},\texttt{H},x))\right],
  \label{eq:harness-objective}
\end{equation}
where $r$ denotes the task-level evaluation function, with the expectation also accounting for stochasticity in agent execution.

\textbf{Inner- and Outer-loop Optimization.}\quad
A standard approach to optimizing \Eqref{eq:harness-objective} is to organize search into an \emph{inner loop} and an \emph{outer loop} \citep{lee2026meta, lee2026recursive, zhang2026self}. The inner loop evaluates a candidate harness by running the corresponding agent on a set of task instances and collecting feedback, including task scores, execution traces, tool outputs, and failures. The outer loop then uses this feedback to diagnose the current harness and propose an improved candidate. Repeating these stages yields an iterative search process in which agent executions provide the evidence used to update the harness.

The outer-loop optimizer may range from a hand-designed update rule~\citep{agrawal2026gepa, yang2024large, yuksekgonul2024textgrad} to an LLM-based agent that automatically modifies prompts, control logic, or the harness program itself~\citep{ursekar2026harnessopt, lee2026meta}. In the latter case, Meta-Harness \citep{lee2026meta}, for example, uses a coding agent to search over harness programs by inspecting previous candidates, evaluation results, and execution traces. However, Meta-Harness produces a globally optimized harness that is applied across future task instances. This can be suboptimal when different instances benefit from different harness strategies. For example, a simple coding task can be solved by a lightweight edit-and-test loop, whereas a difficult task may require specialized tools or more extensive verification.

\section{Our Method}
\label{sec:method}

\begin{figure}[t]
\centering
\includegraphics[width=0.9\linewidth]{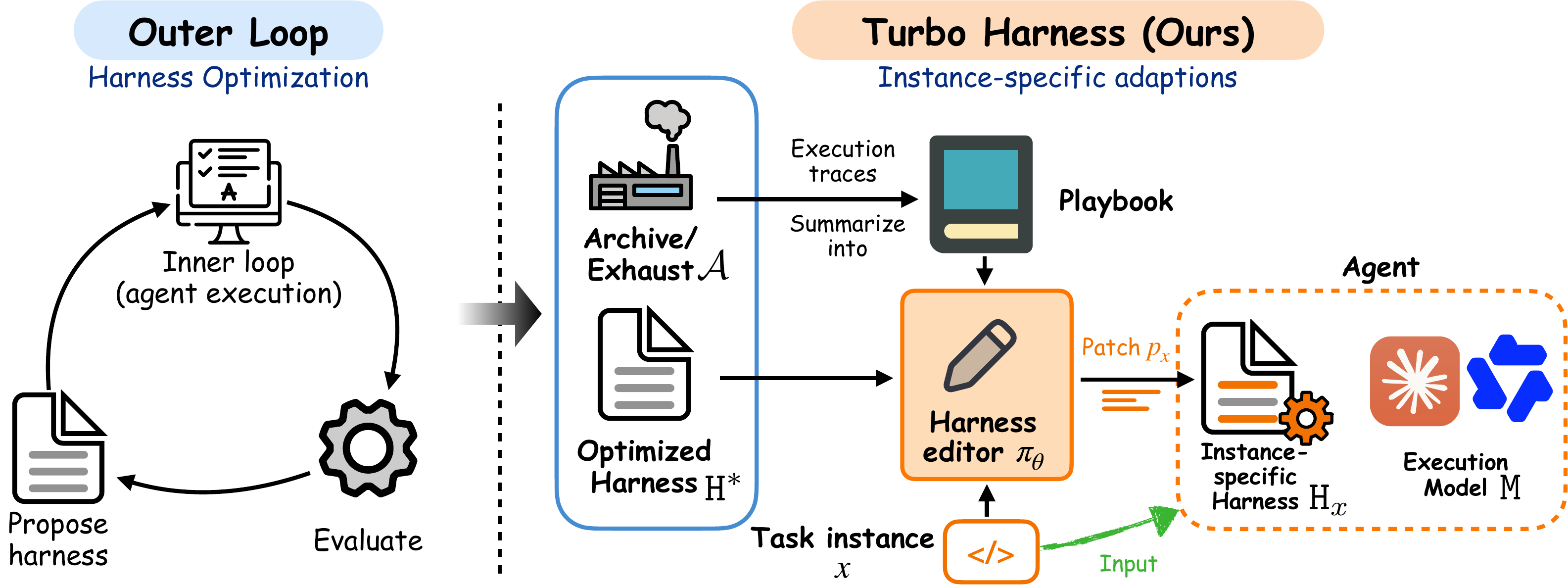}
\caption{\textbf{Turbo Harness overview.} \emph{Left:} an outer loop optimizes a global harness $\texttt{H}^\star$, producing ``exhaust'' (traces, reflections, evaluations) as a byproduct. \emph{Right:} we recycle this exhaust into a \emph{playbook}; a small RL-trained \emph{harness editor} $\pi_\theta$ conditions on an instance $x$ and the playbook to patch $\texttt{H}^\star$ into an instance-specific harness $\texttt{H}_x$, which the frozen model $\texttt{M}$ then runs.}
\label{fig:overview}
\vspace{-2mm}
\end{figure}

We introduce Turbo Harness (Figure~\ref{fig:overview}), which adapts a global harness to each task instance by reusing experience collected during the outer-loop harness search. Our goal is: for each task instance $x$, find an optimal harness $\texttt{H}_x$ such that
\begin{equation}
    \texttt{H}_x^\star \in \arg\max_{\texttt{H}_x \in \mathcal{H}}
  \left[r(\mathsf{Agent}(\texttt{M},\texttt{H}_x,x))\right].
  \label{eq:ins_harness-objective}
\end{equation}
To achieve it, we train a lightweight LLM, which we call the harness editor, to take as input a task instance together with experience gathered during the outer-loop search and propose edits to the global harness. Below, Section~\ref{sec:instance-specific} describes our instance-specific adaptation method and Section~\ref{sec:editor} shows how we train the harness editor.

\subsection{Instance-Specific Harness Optimization}
\label{sec:instance-specific}


\textbf{Harness Editor.}\quad
Searching for an optimal harness independently for each task instance can be expensive, as it may require evaluating multiple candidate harnesses per instance. Instead, we train a harness editor $\pi_\theta$ to propose instance-specific code edits to the globally optimized harness $\texttt{H}^\star$ obtained from the outer-loop optimization in Eq.~\ref{eq:harness-objective}. The harness editor is a small open-weight LLM with trainable parameters $\theta$ (e.g., we use Qwen3.5-9B in our experiments). Because it is lightweight and called only once per task instance, its deployment overhead is minimal.

To further reduce training cost, we reuse the experience collected during outer-loop harness optimization. In particular, the search archive records the outcomes of all evaluated harnesses, including candidates discarded during search. Following ACE~\citep{zhang2026agentic}, we summarize these successes and failures into a \emph{playbook} $\mathcal{P}$ of editing strategies and the conditions under which they are useful. This way, the harness editor does not need to discover new strategies from scratch. It only needs to learn how to use the playbook and identify which previously discovered strategies are relevant to a given task instance. This strategy significantly reduces the training cost and enables the use of a small LLM.

Specifically, given the instance~$x$, the source code of $\texttt{H}^\star$, and the playbook $\mathcal{P}$, the harness editor $\pi_\theta$ generates a patch $p_x$ that produces the adapted harness $\texttt{H}_x$ when applied to $\texttt{H}^\star$:
\begin{equation}
  \pi_\theta(\cdot\mid x,\texttt{H}^\star,\mathcal{P}) \text{ outputs a patch } p_x
  \text{ and let } \texttt{H}_x = \texttt{H}^\star \oplus p_x,
  \label{eq:editor}
\end{equation}
where $\oplus$ denotes applying code edits to a harness. 

\begin{algorithm}[t]
\caption{Turbo Harness}
\label{alg:turbo}
\algrenewcommand{\algorithmicrequire}{\makebox[4.2em][l]{\textbf{Require:}}}
\algrenewcommand{\algorithmicensure}{\makebox[4.2em][l]{\textbf{Output:}}}
\begin{algorithmic}[1]
\Statex \hspace*{-\leftmargin}\textbf{Offline training}\vspace{2pt}
\Require \begin{tabular}[t]{@{}l@{}}
  Outer-loop harness-search archive $\mathcal{A}$; training instances $\mathcal{D}_{\mathrm{train}}$; global harness $\texttt{H}^\star$;\\
  frozen execution model $\texttt{M}$; initial editor $\pi_\theta$; proposals per instance $G$
\end{tabular}
\Ensure Trained editor $\pi_\theta$ and playbook $\mathcal{P}$
\State $\mathcal{P} \gets \Call{Summarize}{\mathcal{A}}$ \Comment{Distill the outer-loop harness-search archive into a playbook}
\For{each training step}
  \State $\mathcal{D}_b \gets \Call{SampleBatch}{\mathcal{D}_{\mathrm{train}}}$ \Comment{Sample a mini-batch of instances}
  \State $\{p_{x,g}\}_{g=1}^G \sim \pi_\theta(\cdot\mid x,\texttt{H}^\star,\mathcal{P}),\quad x\in\mathcal{D}_b$ \Comment{$G$ candidate patches per instance}
  \State $\texttt{H}_{x,g} \gets \texttt{H}^\star \oplus p_{x,g}$ \Comment{Apply each patch to $\texttt{H}^\star$}
  \State $R_{x,g} \gets \Call{Evaluate}{\mathsf{Agent}(\texttt{M},\texttt{H}_{x,g},x)}$ \Comment{Reward from agent execution}
  \State $\theta \gets \Call{GRPO}{\theta,\{(x,p_{x,g},R_{x,g})\}_{x,g}}$ \Comment{Update the editor model (Eq.~\ref{eq:editor-objective})}
\EndFor
\Statex \hspace*{-\leftmargin}\rule{\dimexpr\linewidth+\leftmargin\relax}{0.4pt}
\Statex \hspace*{-\leftmargin}\textbf{Instance-adaptive inference}\vspace{2pt}
\Require \begin{tabular}[t]{@{}l@{}}
  Task instance $x$; global harness $\texttt{H}^\star$; frozen execution model $\texttt{M}$;\\
  trained editor $\pi_\theta$; playbook $\mathcal{P}$
\end{tabular}
\Ensure Task result $y$
\State $p_x \sim \pi_\theta(\cdot\mid x,\texttt{H}^\star,\mathcal{P})$ \Comment{One patch for the test instance}
\State $\texttt{H}_x \gets \texttt{H}^\star \oplus p_x$ \Comment{Apply patch to the global harness}
\State $y \gets \mathsf{Agent}(\texttt{M},\texttt{H}_x,x)$ 
\end{algorithmic}
\end{algorithm}

\vspace{-2mm}


\subsection{Training a Harness Editor}
\label{sec:editor}

We train the harness editor $\pi_\theta$ to favor edits to the global harness $\texttt{H}^\star$ that improve task performance when the frozen LLM executes the task. We optimize the editor with RL, using the resulting task performance as the reward. For each training instance, we evaluate multiple candidate edits and optimize the policy based on their relative rewards.

\textbf{Training Rewards.}\quad
For each instance $x$ in the training set $\mathcal{D}_{\mathrm{train}}$, we sample $G$ candidate harness edits $\{p_{x,g}\}_{g=1}^G$ from $\pi_\theta$. We apply each edit to $\texttt{H}^\star$ to obtain $\texttt{H}_{x,g}$, execute $\mathsf{Agent}(\texttt{M}, \texttt{H}_{x,g}, x)$, and compute a reward $R_{x,g}$ from the resulting task performance and task-specific evaluation metrics. Edits that cannot be applied or fail validation receive zero reward and are not executed.

\textbf{Optimization.}\quad
We optimize $\pi_\theta$ with GRPO~\citep{shao2024deepseekmath}. Let $\mathcal{B}$ be the set of candidate indices in an update minibatch, with $i=(x,g)$ and $p_i=p_{x,g}$. For proposals sampled from the pre-update policy $\pi_{\theta_{\mathrm{old}}}$, the standard GRPO objective is
\begin{equation}
  \mathcal{J}_{\mathrm{GRPO}}(\theta;\mathcal{B})
  = \frac{1}{|\mathcal{B}|}
  \sum_{i\in\mathcal{B}}\frac{1}{|p_i|}\sum_{t=1}^{|p_i|}
  \min\!\left[
    \rho_{i,t}\widehat{a}_i,\,
    \operatorname{clip}\!\left(\rho_{i,t},1-\epsilon,1+\epsilon\right)\widehat{a}_i
  \right]
  -\beta\textnormal{D}_{\scalebox{.6}{\textnormal KL}}\!\left(\pi_\theta\|\pi_{\mathrm{ref}}\right).
  \label{eq:editor-objective}
\end{equation}
Here, $|p_i|$ is the token length of proposal $p_i$, $p_{i,t}$ its $t$-th token, and $p_{i,<t}$ its preceding prefix. The advantage $\widehat{a}_i$ is computed from the rewards $\{R_{x,g}\}_{g=1}^G$ of candidates for the same instance $x$. The token probability ratio is $\rho_{i,t}=\frac{\pi_\theta(p_{i,t}\mid x,\texttt{H}^\star,\mathcal{P},p_{i,<t})}{\pi_{\theta_{\mathrm{old}}}(p_{i,t}\mid x,\texttt{H}^\star,\mathcal{P},p_{i,<t})}$. 

The reference policy $\pi_{\mathrm{ref}}$ is the frozen initial editor, $\textnormal{D}_{\scalebox{.6}{\textnormal KL}}(\pi_\theta\|\pi_{\mathrm{ref}})$ is the sampled KL penalty~\citep{shao2024deepseekmath, ouyang2022training, schulman2017proximal}, and $\epsilon,\beta$ control clipping and KL regularization; the execution model $\texttt{M}$ remains frozen throughout training.

At inference, a single editor call conditions on $\texttt{H}^\star$ and the playbook $\mathcal{P}$ to turn each test instance $x$ into a patch $p_x$, adapting $\texttt{H}^\star$ into an instance-specific harness $\texttt{H}_x$ (falling back to $\texttt{H}^\star$ if the patch fails to apply), which the frozen execution model $\texttt{M}$ then runs; Algorithm~\ref{alg:turbo} and Figure~\ref{fig:overview} give the end-to-end pipeline.


\section{Experiments}
\label{sec:experiments}

We evaluate Turbo Harness on seven benchmarks spanning interactive agent tasks, software engineering, and long-horizon terminal tasks, comparing it with recent harness optimization and prompt-based baselines. We assess gains in task performance and execution efficiency, and conduct ablation studies to investigate where the gains come from.

\subsection{Experimental Setup}
\label{sec:setup}

\textbf{Benchmarks.}\quad
We evaluate Turbo Harness on seven benchmarks spanning agentic tasks, software engineering, and long-horizon terminal interaction. For the four agentic benchmarks, ALFWorld~\citep{shridhar2020alfworld}, ScienceWorld~\citep{wang2022scienceworld}, DBBench~\citep{liu2024agentbench}, and WebShop~\citep{yao2022webshop}, we report success rate, mean environment score in $[0,100]$, accuracy, and Success@1, respectively. For software engineering, we use \emph{SWE-smith-MR}, our multi-repository subset of SWE-smith~\citep{yang2026swe}, and SWE-bench Verified~\citep{jimenez2024swe}. On both coding benchmarks, pass rate is the fraction of test issues resolved by explicitly submitted code patches, with unsubmitted attempts counted as failures. For Terminal-Bench-2.1 (TB2.1)~\citep{merrill2026terminalbench}, we report task pass rate. Higher values indicate better performance for all seven task metrics.

\textbf{Data splits and budgets.}\quad
We evaluate on held-out test instances using the benchmark-specific splits in Appendix~\ref{app:splits}. SWE-smith-MR contains $50$ training and $50$ test issues sampled from $25$ Python repositories. SWE-bench Verified uses a repository-stratified split of $251$ training, $99$ validation, and $150$ test issues. Both coding benchmarks hold out issues within the selected repositories and impose a $40$-step execution limit and a \$3 cost limit per issue. Compared with the $250$-step budget commonly used for SWE-bench Verified evaluation, this tighter budget is intended to stress-test harness quality by limiting the extent to which additional interaction can compensate for ineffective runtime guidance. TB2.1 uses a held-out test split of $44$ tasks.

\textbf{Models.}\quad
We use Qwen3.5-9B~\citep{qwen3.5} for the four agentic benchmarks, evaluate each coding benchmark with Claude Haiku 4.5~\citep{anthropic2025haiku45} and Gemini 3.7 Flash~\citep{google2026gemini37flash} separately, and use Claude Sonnet 4.5~\citep{anthropic2025sonnet45} for TB2.1. Qwen3.5-9B serves as the trainable harness editor in our experiment. Meta-Harness proposals and playbook curation use Claude Sonnet 4.5, except on TB2.1, which use Claude Opus 4.6~\citep{anthropic2026opus46}.

\textbf{Baselines.}\quad
Our primary comparison is \emph{Meta-Harness}~\citep{lee2026meta}, whose globally optimized harness and search archive provide the starting point for Turbo Harness. This comparison evaluates the added value of instance-specific adaptation. We also compare with \emph{Default}, the initial scaffold without harness optimization: mini-swe-agent~\citep{yang2024sweagent} for the coding benchmarks and a basic tool-calling loop for each agentic benchmark. Terminus-Kira~\citep{terminuskira2026} and Terminus-2~\citep{merrill2026terminalbench} serve as the default harnesses on TB2.1. Additional agentic baselines include ReAct~\citep{yao2022react}, Self-Refine~\citep{madaan2023self}, Reflection~\citep{shinn2023reflexion}, and \emph{Harness-R1}~\citep{shao2026harness}. We adapt Harness-R1 to ScienceWorld by porting its code-hook method into the Harness-R1 runtime. Appendix~\ref{app:baselines} details the Default and prompt-based implementations, including Reflection's second attempt; Appendix~\ref{app:harnessr1} documents Harness-R1's runtime and rollout-budget differences.

\textbf{Editor training.}\quad
For each benchmark and execution model, we construct a playbook from the completed Meta-Harness search and hold it fixed during editor training and evaluation. We train the editor with GRPO~\citep{shao2024deepseekmath} using the task-specific rewards in Appendix~\ref{app:reward}. Appendix~\ref{app:checkpoint} specifies checkpoint selection and the WebShop-specific training configuration.

\subsection{Task Performance}
\label{sec:results}

\textbf{Agentic benchmarks.}\quad
Turbo Harness improves over Meta-Harness on all four agentic benchmarks (Table~\ref{tab:agentic}): by $10.0 \%$ on ALFWorld, $7.3$ on ScienceWorld's $[0,100]$ score, $4.2 \%$ on DBBench, and $2.0\%$ on WebShop. It also surpasses the three prompt-based baselines (ReAct, Self-Refine, and Reflection) on every benchmark, and outperforms Harness-R1 on ALFWorld, ScienceWorld, and DBBench while performing comparably on WebShop ($42.0\%$ versus $42.2\%$).

\begin{table}[t]
\centering
\caption{\textbf{Task performance on four agentic benchmarks} (frozen Qwen3.5-9B). Higher is better; metrics are success rate (\%, ALFWorld), mean score ($/100$, ScienceWorld), accuracy (\%, DBBench), and Success@1 (\%, WebShop), with \emph{Avg.}\ their equal-weight mean. Harness-R1 uses two rollouts ($\dagger$: author-reported). Bold denotes the best in each column.}
\label{tab:agentic}
\vspace{0.5em}
\resizebox{\linewidth}{!}{%
\setlength{\tabcolsep}{10pt}
\begin{tabular}{lccccc}
\toprule
Method & ALFWorld & ScienceWorld & DBBench & WebShop & Avg. \\
\midrule
Default               & 40.7 & 25.7 & 57.5 & 34.5 & 39.6 \\
\midrule
\multicolumn{6}{l}{\textit{Prompt-based methods}} \\
ReAct                 & 60.7 & 26.7 & 55.8 & 33.5 & 44.2 \\
Self-Refine           & 40.7 & 21.3 & 53.3 & 18.5 & 33.5 \\
Reflection            & 50.0 & 38.2 & 60.8 & 41.0 & 47.5 \\
\midrule
\multicolumn{6}{l}{\textit{Harness optimization}} \\
Harness-R1& 54.0 & 28.2 & 60.0 & $\mathbf{42.2}^{\dagger}$ & 46.1 \\
Meta-Harness       & 60.7 & 35.1 & 65.0 & 40.0 & 50.2 \\
\rowcolor{oursrow}
\textbf{Turbo Harness (ours)} & \textbf{70.7} & \textbf{42.4} & \textbf{69.2} & 42.0 & \textbf{56.1} \\
\bottomrule
\end{tabular}
}
\vspace{-1em}
\end{table}

\textbf{SWE-smith-MR.}\quad
On SWE-smith-MR, instance-specific adaptation produces large gains with both execution models (Table~\ref{tab:coding}): Turbo Harness raises the pass rate from $50.7\%$ to $64.0\%$ with Haiku and from $70.7\%$ to $88.0\%$ with Gemini, improvements of $13.3$ and $17.3$ percentage points over Meta-Harness. The remaining baselines isolate the harness as the operative lever: averaged across both coding benchmarks, reflective prompt optimization (GEPA)~\citep{agrawal2026gepa} and context engineering (ACE)~\citep{zhang2026agentic} improve over Default but do not reach Meta-Harness ($45.9$ and $44.3$ versus $54.1$), let alone Turbo Harness ($66.4$). A single global harness thus leaves substantial headroom across issues from different repositories, even under a fixed execution model, and adapting the harness per instance recovers much of it.

\textbf{SWE-bench Verified.}\quad
The same pattern holds on SWE-bench Verified, a second coding benchmark drawn from real-world GitHub issues, where Turbo Harness again outperforms or matches Meta-Harness.
With Gemini, the pass rate rises from $38.4\%$ to $54.4\%$; with Haiku, it rises from $56.7\%$ to $59.3\%$, a smaller $2.7$-point improvement.
The size of the gain tracks the executor's remaining headroom: Meta-Harness already lifts Haiku from $31.6\%$ to $56.7\%$, capturing most of the improvement over the default harness and leaving little for instance-specific adaptation, whereas Gemini's lower Meta-Harness baseline leaves ample room that Turbo Harness exploits.


\begin{table}[t]
\centering
\caption{\textbf{Issue resolution and execution efficiency on two coding benchmarks} (frozen Haiku 4.5 and Gemini 3.7 Flash). Per benchmark: pass rate (\%, higher is better), average steps, and cost per issue (both lower is better), as mean $\pm$ standard error of the mean (SEM) over three runs under a $40$-step, \$3 budget. Verified is our held-out $150$-issue subset. Bold marks the best in each column.}
\label{tab:coding}
\vspace{0.5em}
\resizebox{\linewidth}{!}{%
\setlength{\tabcolsep}{5pt}
\begin{tabular}{lcccccc}
\toprule
\multirow{2}{*}{Method} & \multicolumn{3}{c}{SWE-smith-MR} & \multicolumn{3}{c}{SWE-bench Verified} \\
\cmidrule(lr){2-4}\cmidrule(lr){5-7}
 & Pass rate & Steps & Cost (\$) & Pass rate & Steps & Cost (\$) \\
\midrule
\multicolumn{7}{l}{\textit{Execution model: Claude Haiku 4.5}} \\
Default      & $42.0$\,{\small$\pm4.2$} & $29.5$\,{\small$\pm0.8$} & $0.325$\,{\small$\pm0.015$} & $31.6$\,{\small$\pm0.2$} & $35.5$\,{\small$\pm0.2$} & $0.426$\,{\small$\pm0.006$} \\
GEPA         & $44.7$\,{\small$\pm2.9$} & $25.8$\,{\small$\pm0.7$} & $0.308$\,{\small$\pm0.014$} & $41.3$\,{\small$\pm0.4$} & $27.2$\,{\small$\pm1.0$} & $0.352$\,{\small$\pm0.013$} \\
ACE          & $51.3$\,{\small$\pm1.8$} & $26.8$\,{\small$\pm0.4$} & $0.314$\,{\small$\pm0.011$} & $34.7$\,{\small$\pm1.9$} & $35.5$\,{\small$\pm0.4$} & $0.473$\,{\small$\pm0.009$} \\
Meta-Harness & $50.7$\,{\small$\pm2.9$} & $18.5$\,{\small$\pm0.2$} & $0.151$\,{\small$\pm0.004$} & $56.7$\,{\small$\pm2.0$} & $\mathbf{18.8}$\,{\small$\pm0.5$} & $\mathbf{0.153}$\,{\small$\pm0.006$} \\
\rowcolor{oursrow}
\textbf{Turbo (ours)} & $\mathbf{64.0}$\,{\small$\pm1.2$} & $\mathbf{17.3}$\,{\small$\pm0.5$} & $\mathbf{0.136}$\,{\small$\pm0.005$} & $\mathbf{59.3}$\,{\small$\pm1.4$} & $20.4$\,{\small$\pm0.4$} & $0.179$\,{\small$\pm0.006$} \\
\midrule
\multicolumn{7}{l}{\textit{Execution model: Gemini 3.7 Flash}} \\
Default      & $44.0$\,{\small$\pm5.0$} & $26.6$\,{\small$\pm0.9$} & $0.373$\,{\small$\pm0.016$} & $24.4$\,{\small$\pm0.9$} & $33.5$\,{\small$\pm0.2$} & $0.542$\,{\small$\pm0.017$} \\
GEPA         & $57.3$\,{\small$\pm3.5$} & $17.8$\,{\small$\pm0.5$} & $0.213$\,{\small$\pm0.026$} & $40.4$\,{\small$\pm1.1$} & $29.3$\,{\small$\pm0.7$} & $0.433$\,{\small$\pm0.016$} \\
ACE          & $56.7$\,{\small$\pm3.5$} & $21.6$\,{\small$\pm0.2$} & $0.314$\,{\small$\pm0.013$} & $34.7$\,{\small$\pm0.4$} & $33.4$\,{\small$\pm0.3$} & $0.525$\,{\small$\pm0.014$} \\
Meta-Harness & $70.7$\,{\small$\pm2.4$} & $23.1$\,{\small$\pm0.7$} & $0.310$\,{\small$\pm0.015$} & $38.4$\,{\small$\pm0.4$} & $32.5$\,{\small$\pm0.6$} & $0.470$\,{\small$\pm0.012$} \\
\rowcolor{oursrow}
\textbf{Turbo (ours)} & $\mathbf{88.0}$\,{\small$\pm1.2$} & $\mathbf{8.7}$\,{\small$\pm0.4$} & $\mathbf{0.046}$\,{\small$\pm0.004$} & $\mathbf{54.4}$\,{\small$\pm2.3$} & $\mathbf{25.9}$\,{\small$\pm0.3$} & $\mathbf{0.331}$\,{\small$\pm0.010$} \\
\bottomrule
\end{tabular}
}
\vspace{-2mm}
\end{table}

\vspace{-2mm}
\subsection{Execution Efficiency}
\label{sec:efficiency}

Beyond raising accuracy, Turbo Harness also lowers execution cost in three of the four coding settings, resolving each issue in fewer steps and at lower API cost than the global Meta-Harness (Table~\ref{tab:coding}). The savings are largest on SWE-smith-MR with Gemini, where Turbo Harness averages $8.7$ steps per issue rather than $23.1$ and \$0.046 rather than \$0.310, a roughly $6.7\times$ reduction in execution cost; with Haiku it uses $17.3$ steps rather than $18.5$ at \$0.136 rather than \$0.151. On SWE-bench Verified, Gemini shows the same direction, $25.9$ steps rather than $32.5$ and \$0.331 rather than \$0.470.
The exception is Haiku on SWE-bench Verified, the same setting where the pass-rate gain was smallest ($+2.7$ points): there Turbo Harness spends slightly more than Meta-Harness, $20.4$ steps rather than $18.8$ per issue. 
All costs in Table~\ref{tab:coding} cover only the execution model; they exclude both the offline optimization and the one-time per-instance editor call.

\begin{table}[t]
\centering
\caption{\textbf{Terminal-Bench-2.1: performance and efficiency} (frozen Sonnet 4.5, held-out $44$-task test split). Pass rate (\%) is mean $\pm$ SEM over five runs; turns, cost, and input tokens are per-task means. OpenHands, Terminus-Kira, and Terminus-2 are human-designed harnesses; Meta-Harness is automated. Bold marks the best pass rate.}
\label{tab:tb21}
\vspace{0.5em}
\resizebox{\linewidth}{!}{%
\setlength{\tabcolsep}{12pt}
\begin{tabular}{lcccc}
\toprule
Harness & Pass rate (\%) & Turns & Cost (\$/task) & Input tokens \\
\midrule
OpenHands & $47.3$\,{\small$\pm1.5$} & $54.4$ & $0.85$ & $1.55$M \\
Terminus-Kira    & $52.3$\,{\small$\pm1.1$} & $72.2$ & $1.42$ & $2.76$M \\
Terminus-2 & $50.9$\,{\small$\pm0.8$} & $75.2$ & $1.54$ & $3.00$M \\
Meta-Harness         & $50.5$\,{\small$\pm1.7$} & $76.3$ & $1.56$ & $3.15$M \\
\rowcolor{oursrow}
\textbf{Turbo Harness (ours)} & $\mathbf{55.5}$\,{\small$\pm1.4$} & ${71.2}$ & ${1.38}$ & ${2.65}$M \\
\bottomrule
\end{tabular}
}
\end{table}

\subsection{Long-Horizon Terminal Tasks}
\label{sec:tb21}

We evaluate Turbo Harness on Terminal-Bench-2.1 (TB2.1), a benchmark for long-horizon terminal tasks, using a frozen Claude Sonnet 4.5 as the execution model. TB2.1 provides a more challenging test of our premise: unlike on the other benchmarks, the model-optimized harness produced by Meta-Harness does not outperform the human-crafted harnesses (Table~\ref{tab:tb21}). Even so, per-instance adaptation recovers the lead, as Turbo Harness attains the highest mean pass rate of $55.5\%$, $5.0$ points above Meta-Harness and $3.2$ above the strongest default, Terminus-Kira. The gain is consistent, with Turbo Harness posting the top mean in all five runs.

We observe that Turbo Harness lies on the accuracy-cost Pareto frontier. The lightweight OpenHands~\citep{wang2025openhands} harness is the cheapest overall (\$0.85 per task, $54.4$ turns, $1.55$M input tokens) but also the least accurate ($47.3\%$), spending little because it attempts less. Every harness that is competitive on pass rate costs more, yet Turbo Harness is at once the most accurate and the cheapest among them: relative to Meta-Harness it cuts turns per task from $76.3$ to $71.2$, cost from \$1.56 to \$1.38, and input tokens from $3.15$M to $2.65$M, and it likewise undercuts the other competitive harnesses, Terminus-Kira and Terminus-2. Per-instance adaptation thus buys higher accuracy while trimming cost, whereas the heavier global and default harnesses spend more for less.

\subsection{Ablation Studies}
\label{sec:ablation}

\begin{figure}[t]
\centering
\includegraphics[width=0.9\linewidth]{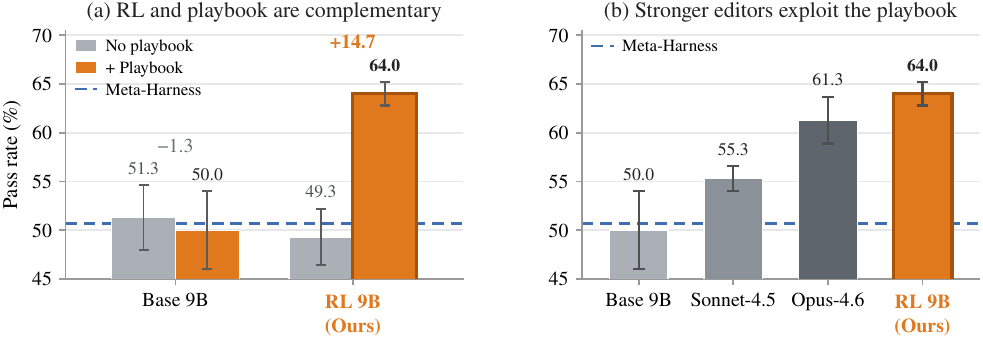}
\vspace{-1em}
\caption{\textbf{Ablation on SWE-smith-MR} (frozen Haiku 4.5; pass rate \%, mean $\pm$ SEM over three runs; dashed line = Meta-Harness). \emph{(a)}~For the lightweight Qwen3.5-9B editor, neither the playbook nor RL alone beats Meta-Harness; combined they reach $64.0\%$. \emph{(b)}~Stronger editors exploit the playbook better, yet our small RL-trained editor matches the best.}
\label{fig:ablation}
\end{figure}

We next study what enables Turbo Harness to improve over the globally optimized harness.
We focus on two questions: whether RL and the recycled playbook are both necessary, and how the editor model's capability and RL training affect the ability to leverage the playbook.

\textbf{RL and the playbook are complementary.}\quad
We first isolate the contributions of RL and playbook conditioning using the same Qwen3.5-9B editor on SWE-smith-MR with a frozen Haiku 4.5 executor (Figure~\ref{fig:ablation} Left).
The globally optimized Meta-Harness achieves a pass rate of 50.7\%.
Introducing an untrained 9B editor without the playbook yields 51.3\%, while providing the same editor with the playbook gives 50.0\%.
Likewise, an RL-trained editor without the playbook reaches only 49.3\%.
In contrast, combining RL training with playbook conditioning raises the pass rate to 64.0\%, a 14.7-point improvement over the same RL-trained editor without the playbook.
Thus, for the lightweight editor, neither generic per-instance editing, access to the playbook alone, nor RL alone explains the gain.
The improvement emerges when RL trains the editor to identify and apply useful strategies distilled from the outer-loop search.

\textbf{RL enables a small editor to exploit the playbook.}\quad
We next vary the editor model while holding the playbook fixed (Figure~\ref{fig:ablation} Right).
The untrained Qwen3.5-9B editor achieves 50.0\%, indicating that simply exposing a small model to the playbook is insufficient.
A stronger frozen Sonnet-4.5 editor improves to 55.3\%, while Opus-4.6 reaches 61.3\%, showing that sufficiently capable models can better leverage the information stored in the playbook even without task-specific fine-tuning. 
Using our RL-trained Qwen3.5-9B editor enables the agent to achieve a 64.0\% pass rate, matching the performance of Opus-4.6 as the harness editor and outperforming Sonnet-4.5, while remaining small enough to run efficiently on local hardware. 
These results suggest that the outer-loop search archive already contains useful harness editing strategies, while RL enables a lightweight model to select and apply them effectively.

\section{Conclusion, Limitations, and Future Work}

We introduced Turbo Harness, an instance-adaptive harness optimization framework that reuses experience from a completed global harness search. Turbo Harness summarizes this search experience into a playbook and trains a lightweight editor to adapt the global harness to each instance. Across our experiments, these instance-specific adaptations consistently outperform the corresponding global harness. Moreover, it provides more flexibility, which we find often translates into lower execution costs.

Our method has two main limitations. First, Turbo Harness builds on a completed global harness search, so its effectiveness depends on the quality and diversity of the harness candidates, trajectories, and feedback collected during that process. If the original search explores only weak strategies or provides limited useful experience, the gains from instance-specific adaptation may also be limited. Second, training the harness editor requires additional environment rollouts. This can be costly on agentic tasks. 
Future work could reduce this training cost through more sample-efficient optimization or reuse of existing rollout data, and investigate whether editors and playbooks can transfer across tasks, domains, or harness-search runs. More broadly, Turbo Harness suggests that the intermediate artifacts produced during harness optimization can serve as reusable experience rather than disposable byproducts.

\subsubsection*{Acknowledgments}
We thank Akash Srivastava for valuable discussions and for sharing relevant literature with us. We are also grateful to Haizhou Shi for taking the time to discuss this work and for the
insights he shared. Finally, we thank members of the Red Hat AI Innovation team
for their help and support.

\bibliography{iclr2027_conference}
\bibliographystyle{iclr2027_conference}

\newpage
\appendix
\section{Benchmarks and Data Splits}
\label{app:benchmarks}

\subsection{Benchmark descriptions}
\textbf{Benchmark descriptions.}\quad
Our seven benchmarks span three families that differ in interaction modality, horizon, and how much the harness can influence behavior.
\begin{itemize}[leftmargin=1.5em]
\item \textbf{ALFWorld}~\citep{shridhar2020alfworld} is a text-based embodied environment in which the agent completes household tasks (e.g., pick-and-place, clean, heat, cool, look, and pick-two) by issuing textual actions and observing environment feedback. Success is a binary indicator of whether the goal configuration is achieved.
\item \textbf{ScienceWorld}~\citep{wang2022scienceworld} is an interactive text environment covering many elementary-science task types (e.g., changes of state, electrical conductivity, and life stages), each requiring a multi-step experimental procedure. The environment returns a partial-credit score in $[0,100]$, and progress is gated by a \texttt{focus} action, making it sensitive to procedure and timing.
\item \textbf{DBBench}~\citep{liu2024agentbench} is a database-interaction benchmark in which the agent answers a question about a relational database by issuing SQL queries and then committing a final answer. Accuracy is measured against the ground-truth answer; we run it against a MySQL backend.
\item \textbf{WebShop}~\citep{yao2022webshop} is a simulated e-commerce environment in which the agent searches and navigates product pages to purchase an item matching a natural-language instruction. The environment returns a dense attribute-match score in $[0,1]$; we report Success@1, the fraction of goals attaining a score of $1$.
\item \textbf{SWE-smith-MR} is our multi-repository subset of SWE-smith~\citep{yang2026swe}. Each instance is a bug-fix issue in a Python repository with a hidden test suite; the agent edits the repository and submits a code patch, which resolves the issue if the associated tests pass. Its construction and rationale are described in Appendix~\ref{app:splits}.
\item \textbf{SWE-bench Verified}~\citep{jimenez2024swe} is a human-validated set of $500$ real-world GitHub issues with associated test suites. As in SWE-smith-MR, an issue is resolved when the agent's submitted patch passes the hidden tests.
\item \textbf{Terminal-Bench-2.1 (TB2.1)}~\citep{merrill2026terminalbench} is a benchmark of long-horizon terminal tasks in which the agent operates a shell to accomplish goals verified by task-specific test scripts. It provides live terminal feedback over long interaction horizons, making the harness an operative lever.
\end{itemize}

\subsection{Data splits}
\label{app:splits}
\textbf{Data splits.}\quad
We use the following benchmark-specific splits:
\begin{itemize}[leftmargin=1.5em]
\item \textbf{ALFWorld:} $100$ training games, $169$ loadable validation games from \texttt{valid\_seen}, and $150$ loadable test games from \texttt{valid\_unseen}.
\item \textbf{ScienceWorld:} $120$ training instances, $60$ validation instances from held-out development variations, and $149$ test instances.
\item \textbf{DBBench:} a stratified split of $300$ tasks into $120$ training, $60$ validation, and $120$ test instances, using seed $42$.
\item \textbf{WebShop:} training goals $1000$--$1119$, validation goals $1500$--$1559$, and $200$ test goals ($0$--$199$), with \texttt{goal\_seed}$=233$. The reported editor is trained on a filtered subset of the training goals, as described below.
\item \textbf{SWE-smith-MR:} we construct this subset from the SWE-smith training split by selecting $25$ Python repositories and sampling two training and two test issues per repository with seed $42$, yielding $50$ training and $50$ test issues. The issue sets are disjoint, but share the same repositories; this evaluates held-out issues within the selected repositories, not generalization to unseen repositories. We do not use a validation set for this subset.
\item \textbf{SWE-bench Verified:} $251$ training, $99$ validation, and $150$ test issues, using a repository-stratified split with seed $42$. The three issue sets are disjoint and together cover all $500$ issues. Meta-Harness search and editor training both use the full $251$-issue training split; the playbook is derived from the training search archive. This is an issue-level split, not a held-out-repository evaluation.
\item \textbf{Terminal-Bench-2.1:} a difficulty-stratified split of $45$ training and $44$ test tasks. Harness search, playbook construction, and editor training use only the training split.
\end{itemize}

\subsection{SWE-smith-MR construction and rationale}
\textbf{SWE-smith-MR design rationale.}\quad
We construct this subset to evaluate instance-specific adaptation across issues with potentially different execution requirements. The selected repositories span web frameworks, data processing, developer tools, and other Python projects, whose project structures and testing conventions may call for different exploration and validation strategies. This motivates our hypothesis that a single global harness can make tradeoffs across repositories that instance-specific adaptation can help resolve. Sampling the same number of issues from each repository gives each repository equal weight in the evaluation, while holding out issues within these repositories tests adaptation to new issues.

\section{Training and Evaluation Protocol}
\label{app:protocol}

\subsection{Playbook construction}
\textbf{Playbook construction.}\quad
The playbook $\mathcal{P}$ is built once per benchmark and execution model from the completed Meta-Harness search archive, following ACE~\citep{zhang2026agentic}, and is then held fixed during editor training and inference. Construction proceeds in three offline stages.

First, \emph{extraction} assembles one experience record per training instance from the search archive, without any model calls. Each record collects the per-harness outcomes and full execution traces recorded during search and classifies the instance as \emph{contrastive} (some harness variant solved it while another failed), \emph{uniformly solved}, or \emph{uniformly unsolved}. For contrastive instances, we additionally pair the most dissimilar solving and failing variants and compute the source-level diff between them, which localizes the harness change responsible for the differing outcome.

Second, \emph{reflection} uses a frontier model to analyze each record and emit a short structured reflection: the salient instance characteristics, the behavioral difference between the harness variants, a candidate editing strategy, a rationale, and a confidence score. Contrastive records are analyzed against both traces and the harness diff, whereas uniform records are analyzed to explain why the outcome did not depend on the harness. Reflections that name known anti-patterns or fall below a confidence threshold are discarded.

Third, \emph{curation} consolidates the surviving reflections into the playbook with a single frontier-model call. Semantically equivalent strategies are clustered into one canonical entry that merges their applicability conditions and aggregates helpful and harmful evidence counts and confidence; strategies whose evidence is net-harmful become \emph{anti-patterns}, and vague or over-specified strategies are filtered out. Each retained entry records the condition under which it applies, the strategy text, its evidence and confidence, and a type tag distinguishing edits to the scaffold's control flow (\textsc{scaffold\_modification}) from edits to the information shown to the model (\textsc{text\_injection}). The result is a compact list of conditional strategies together with a list of anti-patterns and a few general notes; for example, the TB2.1 playbook comprises six strategies and nine anti-patterns. At both training and inference time, the entire playbook is placed in the editor's context, so the editor learns which previously discovered strategies are relevant to each instance rather than retrieving them.

\subsection{Harness editor training}
\textbf{Training configuration.}\quad
All harness editors are Qwen3.5-9B models, full-parameter fine-tuned with FSDP (no low-rank adaptation) using GRPO. Unless noted below, we use a learning rate of $10^{-6}$, a KL-penalty coefficient $\beta=10^{-3}$, within-group advantage standardization, one gradient update per batch with a per-GPU micro-batch of $1$, and rollout sampling at temperature $0.8$ (top-$p$~$0.999$). Rollouts are served with vLLM at tensor-parallel size $1$, and generations that do not terminate normally receive zero reward. We checkpoint every $10$ steps ($5$ for TB2.1). At inference, the editor decodes with temperature $0.3$ and up to $16{,}384$ tokens. Table~\ref{tab:hparams} lists the per-benchmark settings, where the group size $G$ is the number of candidate edits sampled per instance. The abstention penalty and reward-shaping terms are described in Appendix~\ref{app:reward}, and checkpoint selection in Appendix~\ref{app:checkpoint}.

\begin{table}[t]
\centering
\caption{\textbf{Harness-editor GRPO training settings per benchmark.} All editors are Qwen3.5-9B, full-parameter fine-tuned; $G$ is the group size (candidate edits per instance). WebShop uses the tuned recipe (see below): it disables within-group standardization (Dr.\ GRPO), an improvement bonus, and an abstention penalty.}
\label{tab:hparams}
\vspace{0.5em}
\resizebox{\linewidth}{!}{%
\setlength{\tabcolsep}{9pt}
\begin{tabular}{lcccccc}
\toprule
Benchmark & Epochs & Batch & Group $G$ & Max prompt & Max gen. & GPUs \\
\midrule
SWE-smith-MR       & $5$ & $4$  & $8$  & $16384$ & $4096$ & $8$ \\
SWE-bench Verified & $5$ & $8$  & $16$ & $16384$ & $4096$ & $8$ \\
ScienceWorld       & $3$ & $8$  & $8$  & $24576$ & $8192$ & $4$ \\
ALFWorld           & $3$ & $8$  & $8$  & $16384$ & $8192$ & $4$ \\
DBBench            & $3$ & $8$  & $8$  & $16384$ & $8192$ & $4$ \\
WebShop            & $8$ & $4$  & $16$ & $16384$ & $8192$ & $4$ \\
Terminal-Bench-2.1 & $5$ & $16$ & $16$ & $32768$ & $8192$ & $8$ \\
\bottomrule
\end{tabular}
}
\end{table}

\textbf{WebShop training.}\quad
The reported WebShop result uses a domain-specific GRPO configuration. We retain the $65$ training goals whose precomputed global-harness scores lie strictly between $0$ and $1$, sample $16$ proposals per instance, and train for eight epochs. We disable division by the within-group reward standard deviation when computing advantages and use the improvement bonus and abstention penalty in Appendix~\ref{app:reward}. This configuration differs from the training recipe used for the other benchmarks.

\subsection{Reward functions}
\label{app:reward}
\textbf{Reward evaluation.}\quad
We evaluate each valid editor proposal with one execution of the frozen model. A valid patch runs the adapted harness $\texttt{H}_x$, whereas an explicit \texttt{NO\_PATCH\_NEEDED} response runs the global harness $\texttt{H}^\star$. Malformed proposals and patches that fail application or validation receive zero reward without an execution. The training rewards below include shaping terms; the reported test metrics exclude these terms.

\textbf{Task rewards.}\quad
Before applying the abstention penalty described below, we compute rewards as follows:
\begin{itemize}[leftmargin=1.5em]
\item \textbf{SWE-smith-MR and SWE-bench Verified:} $1-s/(2L)$ if the submitted code patch resolves the issue, and $0$ otherwise, where $s$ is the number of execution-model steps and $L=40$ is the step limit. This reward favors successful fixes with fewer steps.
\item \textbf{ScienceWorld:} the environment score divided by $100$, with negative scores clamped to zero.
\item \textbf{ALFWorld:} the binary task-success indicator in $\{0,1\}$.
\item \textbf{DBBench:} the binary task-correctness indicator in $\{0,1\}$.
\item \textbf{Terminal-Bench-2.1:} the binary task-success indicator returned by the task verifier.
\item \textbf{WebShop:} the environment's dense score in $[0,1]$, plus a bonus of $0.1$ when a non-abstaining proposal yields a score greater than the precomputed global-harness score for the same training goal. This bonus rewards improvements over the global harness. Evaluation uses Success@1, defined as the fraction of goals that attain an environment score of $1$.
\end{itemize}

\textbf{Abstention penalty.}\quad
On ScienceWorld, ALFWorld, DBBench, WebShop, and TB2.1, we subtract $0.05$ from the reward of an explicit abstention and clamp the result at zero. Thus, if the unpatched harness earns task reward $r$, the abstention reward is $\max(0,r-0.05)$. This selective penalty encourages exploration of edits while retaining abstention as an option. Neither coding benchmark uses an abstention penalty.

\textbf{Reward integrity.}\quad
We compute task outcomes outside the editor-generated harness code. On the four agentic benchmarks, a trusted runner obtains outcomes directly from the environment, while the harness interacts through an interface that omits scoring methods. Source checks also reject references to protected scoring internals. On both coding benchmarks, the harness executes in a separate process and the parent process scores the submitted code patch. These safeguards reduce opportunities for a harness to manipulate its training reward.

\subsection{Checkpoint selection}
\label{app:checkpoint}
\textbf{Checkpoint selection.}\quad
We select editor checkpoints using validation performance on ALFWorld and ScienceWorld. For SWE-smith-MR and DBBench, we report the final checkpoint at step $60$; for the tuned WebShop run, we report the final checkpoint at step $64$. For TB2.1, we report the step-$5$ checkpoint, at which training was stopped. On SWE-bench Verified, we select by pass rate from one execution run on the $99$ validation issues after excluding checkpoints with collapsed policy entropy. This selects step $20$ for Gemini and step $10$ for Haiku. For Haiku, step $60$ has the highest validation pass rate ($52.5\%$), but is excluded because its training policy entropy has fallen to $0.066$, compared with approximately $0.45$--$0.59$ for steps $10$--$50$. Among these remaining checkpoints, step $10$ has the highest validation pass rate ($49.5\%$). This diagnostic exclusion is part of the reported selection rule; selection is not an unrestricted maximum over validation scores.

\subsection{Execution and scoring protocol}
\label{app:execution}
\textbf{Coding execution.}\quad
Both coding benchmarks use the mini-swe-agent scaffold~\citep{yang2024sweagent} in Docker, with a limit of $40$ execution-model steps and a \$3 cost limit per issue. Only explicitly submitted code patches are evaluated for issue resolution, and the pass-rate denominator includes every test issue. For Turbo Harness, we generate one adapted harness per issue and reuse it across the three execution runs. We compute each run's pass rate, mean steps, and mean execution-model cost over the complete test set, then report the mean and sample standard deviation divided by $\sqrt{3}$ across runs. The Verified subset and execution budget differ from full-benchmark leaderboard settings, so our absolute rates should not be read as leaderboard results.

\textbf{Result aggregation.}\quad
We aggregate results over repeated execution runs where available. For the coding benchmarks, we report the mean and standard error of the mean (SEM) over three runs on the same test issues, with Turbo Harness reusing the generated per-instance harnesses across runs. For TB2.1, pass rates and efficiency metrics report the mean and SEM over five runs. These error bars quantify variability across execution runs; variation across editor training seeds and data splits is not evaluated. Each of the four agentic benchmarks uses one evaluation run per test set.

\section{Baseline Details}
\label{app:baselines}

\subsection{Default harnesses}
\textbf{Coding Default.}\quad
For SWE-smith-MR and SWE-bench Verified, Default runs mini-swe-agent's \texttt{DefaultAgent} with the system, task, observation, and format-error templates from its default configuration. The model receives the issue description, generates a shell command, and observes its output in a Docker environment; this loop continues until explicit submission or the execution limit. The prompt instructs the model to inspect the repository, edit the code, and verify the fix. We set the same $40$-step and \$3 execution limits used for the other coding harnesses. No optimized harness or per-instance harness edit is supplied to this baseline.

\textbf{Agentic Defaults.}\quad
Each agentic Default is a fixed, zero-shot tool-calling loop in our task runtime. It initializes the model with the domain's system prompt and task observation, appends actions and environment feedback to the conversation, and re-prompts when the model fails to produce the required tool call. The domain-specific interfaces are:
\begin{itemize}[leftmargin=1.5em]
\item \textbf{ALFWorld:} a \texttt{take\_action} tool, with admissible actions shown at each step and model outputs canonicalized to an admissible action.
\item \textbf{ScienceWorld:} a \texttt{take\_action} tool accepting textual environment commands, with action-template hints in the initial prompt and subsequent observations truncated to $800$ characters.
\item \textbf{DBBench:} \texttt{execute\_sql} and \texttt{commit\_final\_answer} tools, initialized with the question and database schema; committing the answer ends the rollout.
\item \textbf{WebShop:} \texttt{search\_action} and \texttt{click\_action} tools, with the page observation and available actions supplied after each interaction.
\end{itemize}

\subsection{Prompt-based baselines}
\textbf{Prompt-based variants.}\quad
Our ReAct implementation augments the basic tool-calling loop with reasoning--action demonstrations and an explicit instruction to produce a \texttt{Thought} alongside each action. This distinguishes the ReAct row from the zero-shot Default. Self-Refine adds a proposal--critique--revision cycle before executing an action. Reflection runs an initial episode and, if it is not fully successful, generates a verbal reflection to guide a second episode. It reports the better outcome across the two attempts, whereas Default and Turbo Harness each use a single execution rollout. These variants therefore differ in both interaction strategy and model-call budget.

\subsection{Harness-R1 comparison}
\label{app:harnessr1}
\textbf{Harness-R1 comparison.}\quad
We evaluate the released Harness-R1 checkpoint in its AgentBench runtime on our ALFWorld and DBBench test splits, because we could not reconstruct its exact evaluation sets from the public release. For WebShop, we use the authors' reported result on the same seed-$233$, $200$-goal test set. Because ScienceWorld is not one of Harness-R1's benchmarks, we port its code-hook method into the same runtime as a new task and RL-train the released engineer on ScienceWorld with our optimizer; the reported value ($28.2$) is this RL-trained engineer, and the zero-shot released checkpoint scores comparably ($29.7$). Harness-R1 uses an initial rollout followed by a patched rerun, whereas Turbo Harness proposes its patch before a single execution rollout. The comparison therefore uses different runtimes and rollout budgets; it does not isolate the effect of the harness editor alone.

\section{Additional Results and Analyses}
\label{app:results}

\subsection{Reliability across repeated executions}
\label{sec:analysis}
\begin{figure}[t]
\centering
\includegraphics[width=0.4\linewidth]{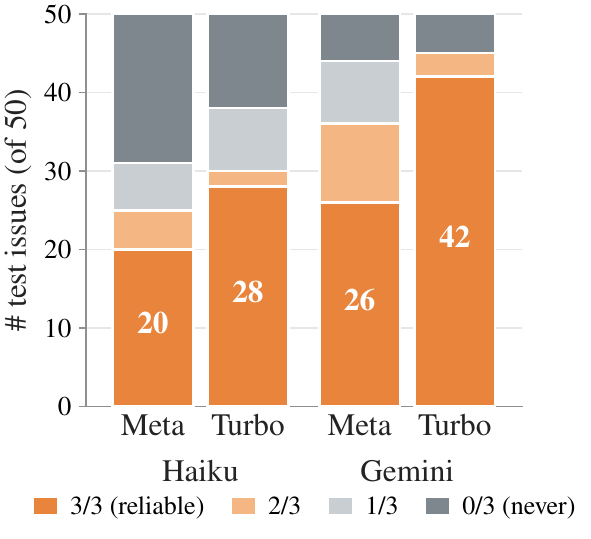}
\caption{\textbf{SWE-smith-MR reliability.} The $50$ test issues grouped by success count over three runs. Dark orange: all three succeed; dark gray: none.}
\label{fig:swe_reliability}
\end{figure}

\textbf{Repeated-execution reliability.}\quad
On SWE-smith-MR, Turbo Harness increases the number of issues resolved in all three runs (Figure~\ref{fig:swe_reliability}). This count rises from $20$ to $28$ out of $50$ issues with Haiku and from $26$ to $42$ with Gemini. At the same time, the number resolved in only one or two runs falls from $11$ to $10$ and from $18$ to $3$, respectively. Thus, the higher mean pass rates are accompanied by more consistently solved issues, especially with Gemini. This analysis measures execution consistency for fixed per-instance harnesses; it does not measure variability in editor training or harness generation.

\subsection{Qualitative case studies of instance-specific patches}
\label{app:cases}
To illustrate what the harness editor actually changes, we show representative patches. The three SWE examples are recovered verbatim from the reported runs and paired with the held-out test outcome; the Terminal-Bench-2.1 example is an outer-loop (search-stage) edit on a training task, included only to illustrate the mechanism because per-instance test patches were not persisted for that benchmark. Diffs are elided (\texttt{...}) but otherwise verbatim.

\textbf{Same global harness, opposite per-instance guidance (SWE-smith-MR).}\quad
On two issues from different libraries, the editor patches the same regions of the \emph{verify-once-submit} $\texttt{H}^\star$ with different guidance, and both flip from failing to fully solved ($3/3$). For a Jinja2 regression it injects a history-inspection step:
\begin{codebox}
  ## Recommended Workflow
  1. Analyze the codebase by finding and reading relevant files.
+    For regression bugs, check git history to understand what
+    changed before fixing.
+ ### Investigate changes (for regression bugs):
+ git log --oneline -20
+ git diff HEAD~1..HEAD --stat
\end{codebox}
whereas an \texttt{arrow} (datetime library) use-before-assignment bug receives unrelated guidance rather than the history-inspection step. The same $\texttt{H}^\star$ is thus specialized differently per instance.

\textbf{Reallocating the step budget (SWE-bench Verified).}\quad
On the \texttt{django-13315} issue, whose $\texttt{H}^\star$ imposes an explicit per-phase step budget, the editor relaxes an aggressive ``submit immediately'' gate to buy exploration and a real fix-and-verify budget:
\begin{codebox}
- ### Phase 1: Understand (5-10 steps)
+ ### Phase 1: Understand (10-15 steps)
- ### Phase 3: Execute Fix (1 step)
+ ### Phase 3: Execute Fix (5-10 steps)
- ### Phase 4: Submit IMMEDIATELY (1 step)
+ ### Phase 4: Finalize and Submit (5-10 steps)
\end{codebox}
The outcome improves from $1/3$ (Meta-Harness) to $3/3$ (Turbo Harness). The budget is instance-specific: on \texttt{sympy-16766} (also $1/3\rightarrow3/3$) the editor enlarges only Phase~3 and \emph{keeps} the one-step submit gate.

\textbf{Editing executable loop code, not prompt wording (SWE-smith-MR).}\quad
On a \texttt{gunicorn} issue, the editor rewrites the scaffold's control code rather than its text. It fires the verification gate one step earlier and teaches the edit detector to also recognize append redirection as a source edit:
\begin{codebox}
  if (not self.verification_triggered and self.has_edited_source
-     and self.step_count >= 8
+     and self.step_count >= 6
      and self._is_verification_command(self.last_action)):
          observation += submit_reminder
  ...
  patterns = [ r'sed\s+.*\.py', r'cat\s+.*>\s*\S+\.py',
+              r'echo\s+.*>>\s*\S+\.py' ]
\end{codebox}
The instance flips from $0/3$ to $3/3$, and the threshold is set per instance ($5$, $6$, or $10$ across different bugs) rather than as a fixed global knob, confirming that adaptation reaches the executable harness and not only the prompt.

\textbf{One knob, opposite optimum per task (Terminal-Bench-2.1).}\quad
The winning $\texttt{H}^\star$ escalates to forced tool use only after two stalled turns. On \texttt{chess-best-move}, the model instead spends turns reasoning in prose, so an edit that forces a tool call every turn solves the task:
\begin{codebox}
  completion_kwargs = {"model": ..., "tools": TOOLS,
+                      "tool_choice": "required",
                       ...}
\end{codebox}
Yet forcing tool use \emph{globally} is a playbook anti-pattern that regresses the whole suite by $6.7\%$, and \texttt{extract-elf} is instead solved by the adaptive escalation and fails under the global forced setting. The same control knob therefore has opposite optima on different tasks, which a single global harness cannot satisfy but instance-specific adaptation can. (These are search-stage edits on training tasks, shown only to illustrate the mechanism.)

\end{document}